\documentclass[letterpaper, 10 pt, conference]{ieeeconf}

\IEEEoverridecommandlockouts
\usepackage{booktabs}
\usepackage{makecell}
\usepackage{multirow}
\usepackage{graphicx}
\usepackage{amsmath}
\usepackage{amssymb}
\usepackage{array}
\usepackage{url}
\usepackage{xcolor}
\usepackage{cuted}
\usepackage{stfloats}
\usepackage[font=small]{caption}
\usepackage{marvosym}
\usepackage{cite}    
\makeatletter
\let\NAT@parse\undefined
\makeatother
\usepackage[colorlinks,linkcolor=red,citecolor=green,urlcolor=blue]{hyperref}
\graphicspath{{images/}}
\usepackage[ruled,vlined,linesnumbered]{algorithm2e}
\title{\LARGE \bf
ParkingWorld: End-to-End Autonomous Parking Reinforcement Learning from Corrective Experience in 3DGS Simulation
}

\author{
Zhengcheng Yu$^{1,\dagger}$\quad 
Changze Li$^{2,\dagger}$\quad
Haoran Liu$^{2}$\quad
Tong Qin$^{2,\text{\Letter}}$\\[0.5em]
$^{1}$ Tsinghua University \quad
$^{2}$ Shanghai Jiaotong University\\[0.5em]
Project Website: \url{https://yu-zhengcheng-11.github.io/ParkingWorld/}
\thanks{$\dagger$ Equal contribution. \quad \text{\Letter} Corresponding author.}
}

\makeatletter
\let\@oldmaketitle\@maketitle
\renewcommand{\@maketitle}{\@oldmaketitle
	\vspace{0.2cm}
	\centering
	\setcounter{figure}{0}
	\begin{minipage}{1.0\linewidth}
		\includegraphics[width=1.0\textwidth]{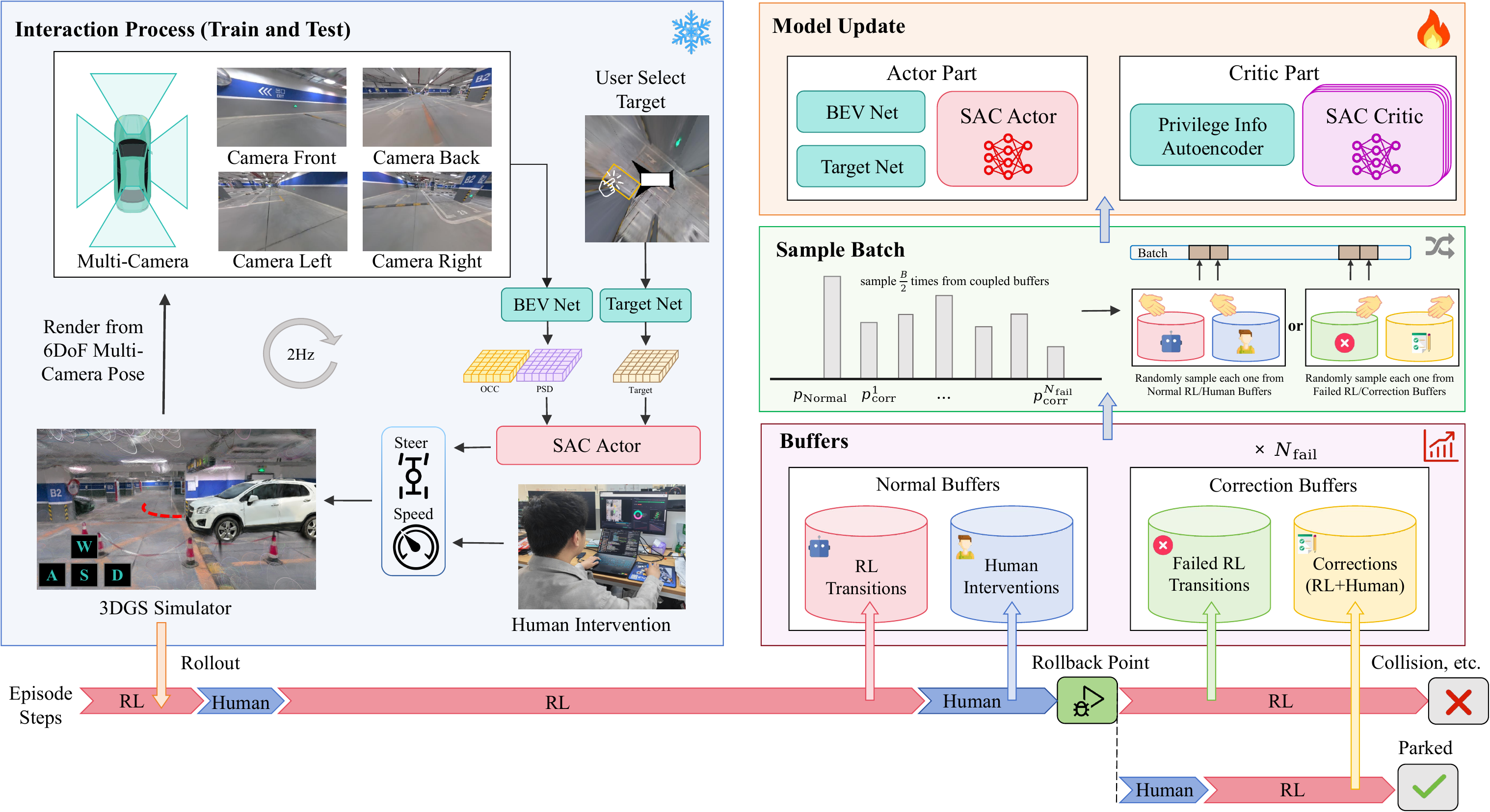}
		\vspace{0cm}
		\captionof{figure}{System pipeline of the proposed human-in-the-loop autonomous parking framework. The left panel shows the interaction during training and testing: multi-camera views rendered from the 3DGS simulator are encoded into BEV and target features, then fed into the SAC actor for low-level steering and speed control which are optionally corrected by human intervention. The right panel summarizes model update: normal RL transitions, human interventions, failed RL transitions, and rollback-based correction trajectories are stored in coupled replay buffers and sampled hierarchically to update the actor and critic. The bottom timeline illustrates how RL rollout, human takeover, and rollback correction are linked within an episode.}
        \label{system_overview}
	\end{minipage}
	\vspace{-0.3cm}
}
\makeatother

\begin{document}

\maketitle
\thispagestyle{empty}
\pagestyle{empty}

\begin{abstract}
Autonomous parking demands precise low-speed maneuvering within narrow, cluttered, and highly constrained environments, where vehicles must navigate tight spaces while avoiding static obstacles and complex geometric boundaries. Unlike imitation learning, which typically requires massive volumes of high-quality expert demonstrations to converge to a stable policy and often suffers from limited generalization to unseen scenarios, traditional reinforcement learning (RL) methods face persistent challenges including excessive training overhead, inefficient exploration, and even failure to learn viable parking strategies in challenging settings. To address these limitations, this paper presents a correction-in-the-loop sample-efficient reinforcement learning (CIL-SERL) framework for end-to-end autonomous parking, which is entirely trained in a photorealistic 3D Gaussian Splatting (3DGS) parking simulator that enables high-fidelity digital reconstruction of real-world scenes. Inspired by error-correction notebooks used in learning practice, we design a novel multi-level replay buffer mechanism. These buffers hierarchically organize and store standard RL rollouts, human corrective interventions, failed exploration trajectories, and rollback-based correction segments in separate yet interconnected memory regions, facilitating structured sampling and targeted learning during training. The proposed framework is systematically evaluated in both the 3DGS simulation environment and a physical vehicle platform. Extensive experimental results demonstrate that our method achieves substantial improvements in parking success rate, operational efficiency, and safety performance across diverse scenarios, validating the effectiveness and practical applicability of the proposed CIL-SERL-based end-to-end autonomous parking solution.
\end{abstract}


\section{Introduction}
\begin{quote}
\itshape
The only real mistake is the one from which we learn nothing.
\par\noindent\rule{\linewidth}{0.4pt}
\par\raggedleft\upshape
Henry Ford
\end{quote}

Autonomous parking is a compact but demanding testbed for robot learning. The vehicle must reason about static obstacles, available free space, parking slot geometry, nonholonomic constraints, and repeated direction changes.

\begin{figure}[t!]
   \centering
   \includegraphics[width=0.5\textwidth]{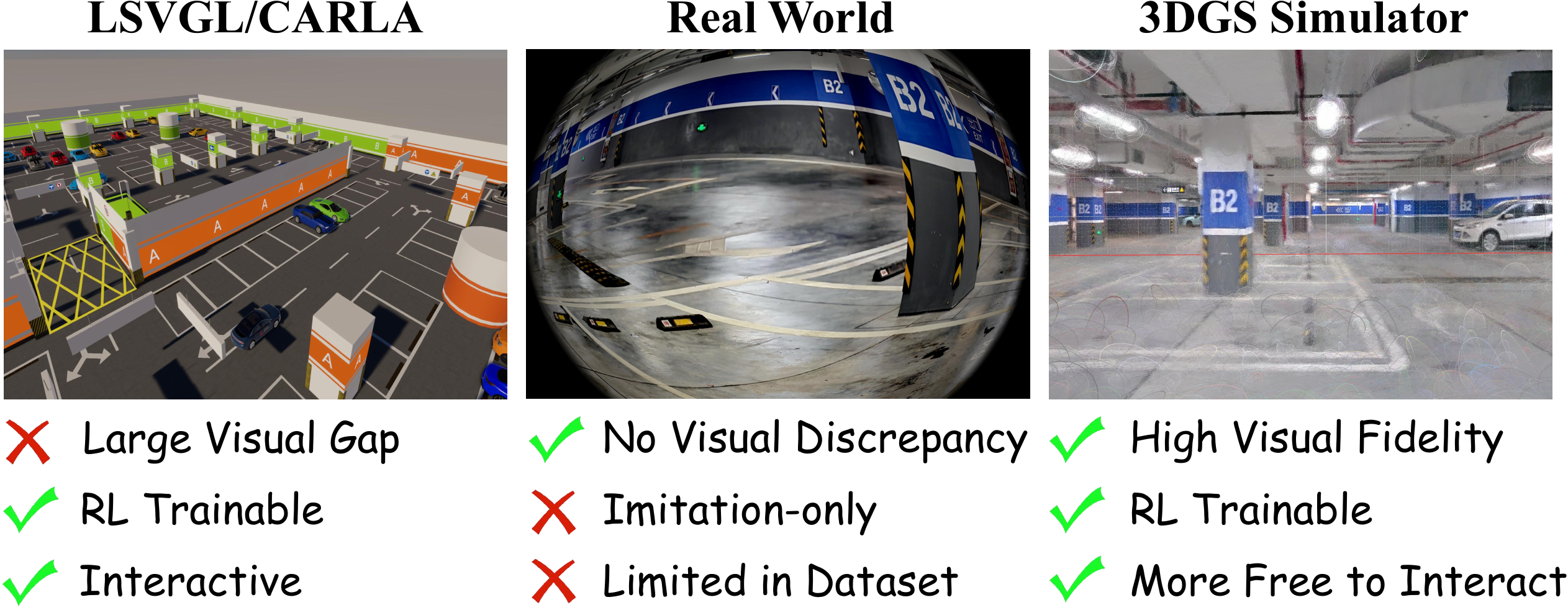}
   \caption{Comparison of training in LGSVL/CARLA simulators, the real world, and our 3DGS simulator.}
   \label{three_ways_comparison}
\end{figure}

Reinforcement learning for autonomous parking is usually implemented in simulators such as CARLA\cite{CARLA} and LGSVL\cite{LGSVL}, which exhibit a significant visual domain gap compared with real-world scenarios. This domain mismatch results in trained policies being unable to be directly deployed to physical vehicles, as Fig.~\ref{three_ways_comparison}. To address this critical issue, we construct a 3D Gaussian Splatting (3DGS) \cite{3dgs} simulator that achieves digital twin reconstruction of real parking scenes. The policies trained in such reconstructed scenarios can achieve zero-shot generalization to real-world parking environments, effectively bridging the gap between simulation training and real-vehicle deployment \cite{reap}. Since the simulator acts as a self-contained virtual space with inherent physical rules and supports human interaction, scene expansion, training, and evaluation, we term our overall autonomous parking learning framework ParkingWorld in this work.

In RL training, we retain the autonomy of an off-policy RL system \cite{sac} while introducing a human-in-the-loop correction mechanism for unproductive or failed exploration. Instead of requiring humans to label all states or replacing reinforcement learning with imitation learning, the proposed framework uses sparse human interventions as corrective experience. When the policy reaches a failure state, the system records both the failed autonomous behavior and the corresponding human correction as a paired experience. These pairs form a structured ``mistake notebook" that reshapes the replay distribution toward informative failure cases, allowing the policy to learn not only from successful exploration but also from how its mistakes are corrected.

The contributions of this paper are summarized as follows:
\begin{itemize}
\item We propose an interactive 3DGS simulator for autonomous parking based on the non-interactive simulator in \cite{reap}. Our simulator not only supports multi-view image rendering synchronized with vehicle motion to provide realistic sensory observations, but also enables interactive operation by human operators, laying a solid foundation for human-in-the-loop training.
\item We design a correction-in-the-loop sample-efficient RL (CIL-SERL) mechanism that records failed autonomous rollouts and successful human-guided corrections as paired replay evidence, enabling the policy to learn from corrected mistakes and reducing repeated ineffective exploration.
\item We conduct extensive closed-loop verification experiments on both the proposed 3DGS simulation platform and a real-world parking testbed. Comprehensive experimental results demonstrate the superior performance of our proposed framework.
\end{itemize}

\section{Literature Review}
\subsection{End-to-End Autonomous Parking}

End-to-end autonomous parking has evolved rapidly in recent years, with research progressing from imitation learning (IL) toward reinforcement learning (RL). Early IL-based methods focused on learning expert parking behaviors directly from demonstrations. Yang et al. proposed E2EParking\cite{e2e-parking}, an imitation learning framework trained in the CARLA simulator, which mapped surround-view images and vehicle states to control signals. Li et al. developed ParkingE2E\cite{parking-e2e}, a Transformer-based end-to-end planner using bird’s-eye-view (BEV) features, which was validated in real parking garages but remained sensitive to distribution shifts.

To overcome the data dependency of imitation learning, reinforcement learning (RL) has been increasingly adopted for autonomous parking. Wu et al. \cite{rled} combined soft actor-critic (SAC) with expert demonstrations, using data augmentation and mixed priority sampling to accelerate training and improve parking performance. Jiang et al. proposed HOPE \cite{hope}, a hybrid policy planner integrating RL with Reeds-Shepp (RS) curves, which leveraged geometric priors and action masking to enhance both training efficiency and robustness in complex environments.

\subsection{Sim-to-Real Transfer for Reinforcement Learning}

Sim-to-real transfer has long been a core challenge in reinforcement learning (RL), as policies trained in simulation often degrade severely when deployed in the physical world. Early work in this field focused on domain randomization to narrow the visual gap between simulation and reality. 

The emergence of 3D Gaussian Splatting (3DGS) revolutionized high-fidelity simulation. Building on 3DGS, RAD\cite{rad} proposed the first large-scale 3DGS-based RL framework for end-to-end driving, combining imitation learning to improve transfer robustness. Meanwhile, Flying in Clutter\cite{flying_3dgs} integrated 3DGS with adversarial domain adaptation, enabling zero-shot transfer for monocular RGB drone navigation in cluttered environments. Zhu et al. proposed VR-Robo\cite{vr-robo}, a real-to-sim-to-real framework that reconstructs photorealistic and physically interactive simulation environments from multi-view RGB
images in the real world using a Gaussian Splatting–mesh
hybrid representation. Most recently, our previous work REAP \cite{reap} further advanced the field by constructing a 3DGS-based parking simulator and proposing a Real2Sim2Real training pipeline. It successfully deployed end-to-end RL parking policies in the real world, demonstrating the effectiveness of 3DGS in bridging the sim-to-real gap for complex autonomous parking tasks.

\subsection{Reinforcement Learning from Human Experience}
Human-in-the-loop reinforcement learning has evolved as a promising direction to address sample inefficiency and safety challenges in real-world RL. 

Recent works have attempted to incorporate human experience into reinforcement learning for autonomous driving. To mitigate the risk of unreliable human guidance, SafeHIL-RL\cite{safety-aware-hitl} introduced a safety-aware shared-control mechanism based on Frenet-based dynamic potential fields and curriculum guidance. More recently, H-DSAC\cite{h-dsac} integrated proxy value propagation with distributional soft actor-critic, enabling human intent to be propagated through a distributional proxy value function for safe and sample-efficient real-world driving policy learning.

In the manipulation field, HIL-SERL\cite{hil-serl} integrated offline demonstrations with real-time human corrections into a unified framework, achieving near-perfect performance on diverse real-world dexterous manipulation tasks within practical training times. $\pi_{0.6}^*$ \cite{pi-06-star} further leveraged human corrective interventions to refine diffusion-based policies and significantly improve generalization across unseen scenarios.

\section{Methodology}
This section presents the overall framework of our proposed autonomous parking system, consisting of three core components. In Section~\ref{method_3dgs_simulator}, we elaborate on the construction and functionalities of the 3D Gaussian Splatting (3DGS) simulator, which serves as a high-fidelity, physically interactive environment for policy training. In Section~\ref{method_e2e_architecture}, we detail the architecture of the end-to-end reinforcement learning autonomous parking network. Finally, in Section~\ref{method_correction_rl}, we describe our correction-in-the-loop RL training method.

\subsection{Interactive 3DGS Autonomous Parking Simulator}
\label{method_3dgs_simulator}

\begin{figure*}[t!]
   \centering
   \includegraphics[width=1\textwidth]{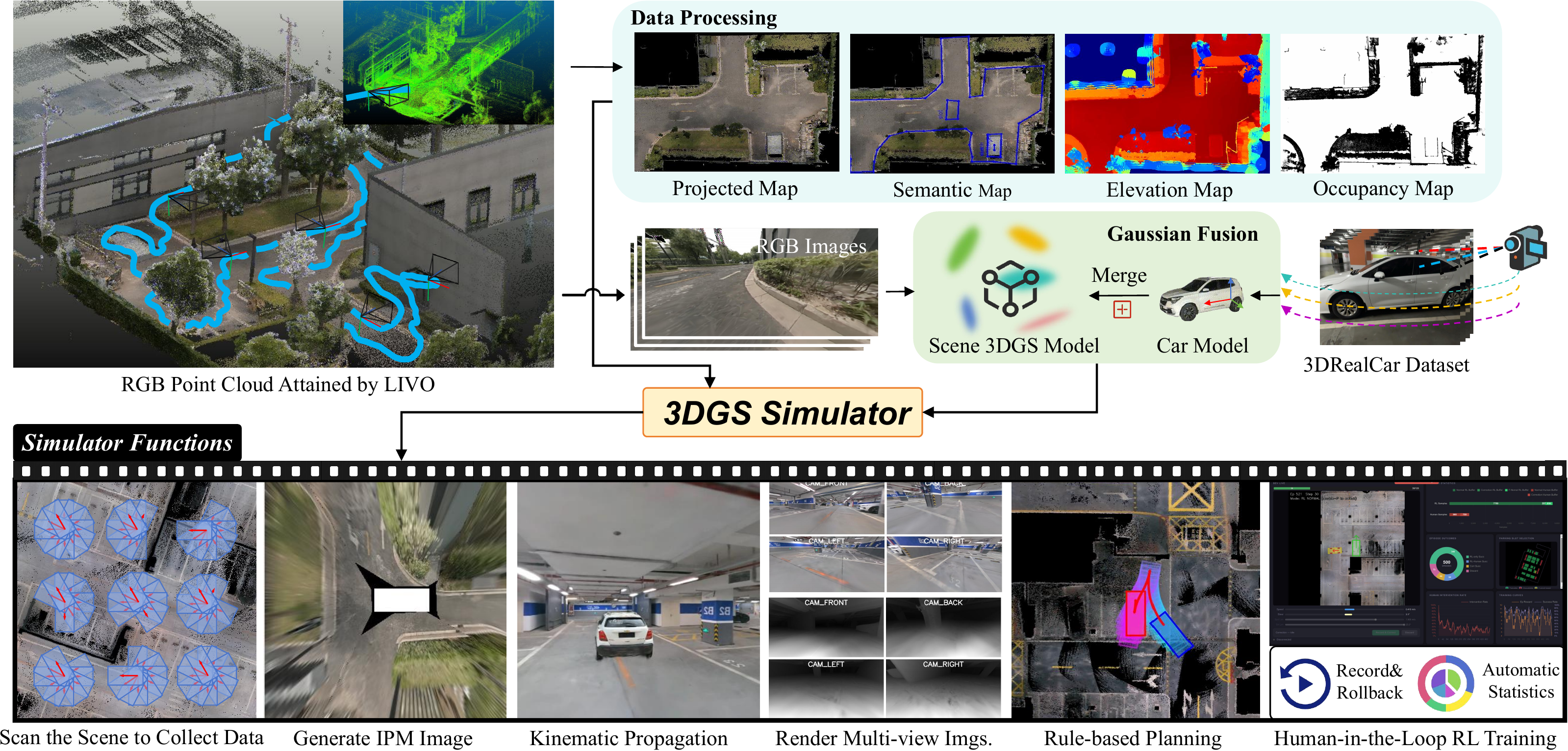}
   \vspace*{-0.5em}
   \caption{Overview of the proposed ROS-based interactive 3DGS simulator for autonomous parking. The simulator integrates LIVO-derived map representations, scene-level and vehicle-level 3DGS models, and supports photorealistic multi-view rendering, vehicle dynamics, collision checking, rule-based planning, and human-in-the-loop RL. It provides synchronized perception, state feedback, rollback recording, and training statistics for closed-loop parking policy development.}
   \label{3dgs_simulator}
\end{figure*}

We construct an interactive 3DGS simulator for autonomous parking by building digital twins of real-world parking scenes following \cite{reap}. The reconstruction process starts from data collected in target environments, such as underground garages, using the XGRIDS Lixel K1 handheld mobile mapping device equipped with IMU, LiDAR, and RGB cameras. The handheld mapping device integrates both the tightly coupled LIVO algorithm \cite{fast-livo2} and the 3DGS reconstruction pipeline. The LIVO module first estimates sensor poses and generates dense colorized point clouds with centimeter-level geometric accuracy. Then, the built-in 3DGS module converts the registered point clouds and associated multi-view RGB images into a scene-level Gaussian representation, where the scene geometry and appearance are encoded by anisotropic Gaussians with learnable position, covariance, opacity, and spherical harmonics coefficients, which are $\left\{
    \left(
    \boldsymbol{\mu}_{i},
    \boldsymbol{\Sigma}_{i},
    \alpha_i,
    \mathbf{c}_i
    \right)\right\}$. Specifically, each Gaussian primitive is represented as 
\begin{equation}
    \mathcal{G}_i(\mathbf{x}) =
    \exp\left(-\frac{1}{2}(\mathbf{x}-\boldsymbol{\mu}_i)^{T}
    \boldsymbol{\Sigma}_i^{-1}
    (\mathbf{x}-\boldsymbol{\mu}_i)\right),
\end{equation}
where $\boldsymbol{\mu}_i$ and $\boldsymbol{\Sigma}_i$ denote its 3D mean and covariance. The covariance is parameterized by a rotation matrix $\mathbf{R}_i$ and a scaling matrix $\mathbf{S}_i$ as $\boldsymbol{\Sigma}_i = \mathbf{R}_i \mathbf{S}_i \mathbf{S}_i^{T} \mathbf{R}_i^{T}$.
For rendering, each 3D Gaussian is projected to the image plane, producing a 2D covariance
\begin{equation}
    \boldsymbol{\Sigma}'_i =
    \mathbf{J}_i \mathbf{W}_i \boldsymbol{\Sigma}_i
    \mathbf{W}_i^{T} \mathbf{J}_i^{T},
\end{equation}
where $\mathbf{W}_i$ is the view transformation and $\mathbf{J}_i$ is the Jacobian of the perspective projection. The final pixel color is obtained by front-to-back alpha compositing:
\begin{equation}
    \mathbf{C}(\mathbf{p}) =
    \sum_{i \in \mathcal{N}(\mathbf{p})}
    T_i \alpha_i(\mathbf{p}) \mathbf{c}_i,
    \quad
    T_i = \prod_{j<i}\left(1-\alpha_j(\mathbf{p})\right),
\end{equation}
where $\mathcal{N}(\mathbf{p})$ denotes the set of projected Gaussians that overlap pixel $\mathbf{p}$, $\mathbf{c}_i$ is the view-dependent color decoded from spherical harmonics coefficients, and $\alpha_i(\mathbf{p})$ is determined by the projected Gaussian opacity. The resulting 3DGS model provides a photorealistic digital twin of the parking environment for subsequent simulation and policy training. In parallel, vehicle-level 3DGS models $\mathcal{G}^{v}$ from the 3DRealCar \cite{3d-real-car} dataset are fused with the reconstructed parking scenes $\mathcal{G}^{s}$, allowing the simulator to represent dynamic motion, i.e., $\mathcal{G}=\mathcal{G}^{s}\cup\mathcal{G}^{v\rightarrow s}$, where $\mathcal{G}^{v\rightarrow s}$ denotes vehicle Gaussians transformed into the scene coordinate frame. 

The resulting simulator provides both photorealistic rendering and physically interactive training signals. Given the current vehicle state and control action, the simulator propagates vehicle motion using a kinematic model, updates the camera viewpoints, and renders synchronized multi-view RGB observations for the policy network. It also supports collision detection, elevation-aware state update, rule-based path planning, human-in-the-loop intervention, state rollback recording, and training-statistics logging. Therefore, the simulator serves not only as a visual renderer, but also as a closed-loop autonomous parking environment in which reinforcement learning policies can be trained, corrected, and evaluated under realistic spatial constraints, as Fig.~\ref{3dgs_simulator}.

\subsection{Architecture of End-to-End Reinforcement Learning Autonomous Parking Network}
\label{method_e2e_architecture}
\begin{figure*}[t!]
   \centering
   \includegraphics[width=1\textwidth]{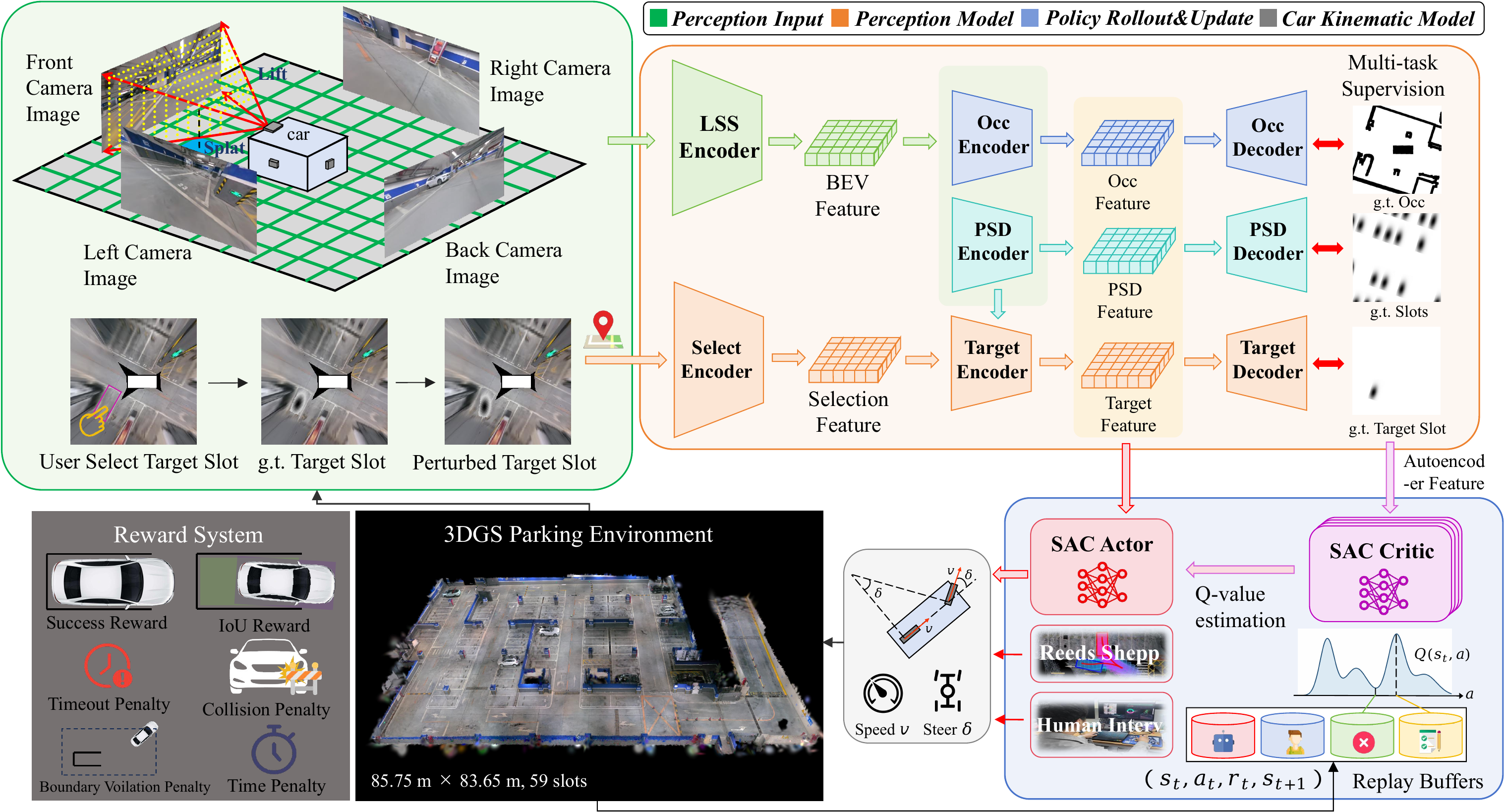}
   \vspace*{-0.5em}
   \caption{Detailed architecture of the proposed end-to-end RL framework for autonomous parking (supplementary to Fig. 1). Multi-view images and perturbed target slots from the 3DGS simulator are encoded into BEV, occupancy, and parking-slot features via a multi-task lift-splat perception model. These features feed into an SAC actor to output vehicle control commands, with optional human correction during training. The single-track kinematic model simulates vehicle motion, while rewards are defined by parking success, IoU, collision, and time efficiency. Transitions are stored in replay buffers to optimize the SAC actor and critic networks.}
   \label{system_detail}
\end{figure*}

The proposed autonomous parking policy follows an end-to-end architecture as Fig.~\ref{system_detail}. At each time step, the 3DGS simulator renders synchronized multi-view RGB images $\mathbf{I}_t=\{I_t^k\}_{k=1}^{N_c}$ according to the current vehicle pose. Together with the manually selected and perturbed target parking slot $\mathbf{s}_t^{tar}$, these observations are fed into a BEV perception module to extract compact scene representations for downstream policy learning. The perception network follows the classical Lift-Splat-Shoot (LSS) \cite{lss} paradigm, where image features from different camera views are lifted into a 3D frustum space and splatted onto a unified BEV plane, yielding the BEV feature $\mathbf{F}_t^{bev}=E_{LSS}(\mathbf{I}_t)$. The BEV feature is further encoded by task-specific heads into occupancy features $\mathbf{F}_t^{occ}=H_{occ}(\mathbf{F}_t^{bev})$, global parking-slot features $\mathbf{F}_t^{slot}=H_{slot}(\mathbf{F}_t^{bev})$, and target-slot features $\mathbf{F}_t^{tar}=H_{tar}(\mathbf{F}_t^{bev},\tilde{\mathbf{F}}_t^{tar})$. These feature representations are supervised by the corresponding perception ground-truth labels in $[-10\mathrm{m}, 10\mathrm{m}]\times[-10\mathrm{m}, 10\mathrm{m}]$, enabling the policy to access structured information about free space, obstacle layout, parking-slot geometry, and the selected target slot.

The extracted multi-modal features are then passed to the SAC-based reinforcement learning module. For the actor, we concatenate the perception features into the policy feature $\mathbf{z}_t^{\pi}=[\mathbf{F}_t^{bev},\mathbf{F}_t^{occ},\mathbf{F}_t^{slot},\mathbf{F}_t^{tar}]$, and the actor network predicts the continuous parking action $\mathbf{a}_t=\pi_{\theta}(\mathbf{z}_t^{\pi})=[\delta_t,v_t]$, where $\delta_t$ and $v_t$ denote the steering angle and signed velocity, respectively. Different from the actor, the critic network uses privileged simulator information through an auxiliary autoencoder branch. Let $\mathbf{G}_t$ denote the simulator-provided ground-truth representation, including structured BEV, occupancy, parking-slot, and target information. The autoencoder encodes it into a latent critic feature $\mathbf{z}_t^{q}=E_{\psi}(\mathbf{G}_t)$ and reconstructs the input as $\hat{\mathbf{G}}_t=D_{\psi}(\mathbf{z}_t^{q})$. Meanwhile, the action is embedded as $\mathbf{z}_t^{a}=\phi_a(\mathbf{a}_t)$. The critic therefore estimates the soft Q-value from the joint state-action feature, i.e., $Q_{\omega_i}(\mathbf{z}_t^{q},\mathbf{z}_t^{a})$ for the twin critics $i\in\{1,2\}$. This design allows the actor to rely on perception-derived features available during policy execution, while the critic benefits from richer privileged simulator supervision for more stable value learning. The predicted action is finally executed through a single-track kinematic vehicle model, producing the next vehicle state, updated observations, and reward signals.

To improve training stability, layer normalization is introduced into the policy and value networks. Let $h \in \mathbb{R}^{d}$ denote an intermediate feature vector before the final value prediction layer, and let
\begin{equation}
    \hat{h} = \mathrm{LN}(h)
\end{equation}
be the normalized feature. Since layer normalization controls the feature scale, the final linear Q-value prediction

\begin{equation}
    Q(s,a) = w^{\top}\hat{h} + b
\end{equation}
is bounded by
\begin{equation}
    |Q(s,a)| \leq \|w\|_2 \|\hat{h}\|_2 + |b|.
\end{equation}
Thus, the Q-value magnitude is bounded by the final-layer weight norm and normalized feature scale. This is particularly important in our human-in-the-loop setting, where abundant high-quality intervention samples may otherwise cause critic overestimation and propagate overly optimistic values to nearby out-of-distribution actions~\cite{rlpd}.

\subsection{Correction-in-the-Loop Sample-Efficient Reinforcement Learning}
\label{method_correction_rl}

The above perception and SAC architecture are trained with a correction-in-the-loop sample-efficient reinforcement learning (CIL-SERL) mechanism. Fig.~\ref{system_overview} and Algorithm 1 illustrate the CIL-SERL training procedure. Different from ordinary human-in-the-loop RL that simply mixes all human interventions into one demonstration buffer, CIL-SERL explicitly couples what the autonomous policy did wrong with how the failure was corrected. We model the parking simulator as an MDP $\mathcal{M}=(\mathcal{S},\mathcal{A},P,r,\gamma)$, where the policy observes $\mathbf{s}_t$, outputs a continuous control action $\mathbf{a}_t=[\delta_t,v_t]$, receives reward $r_t$ (see Appendix A), and generates a transition
\begin{equation}
    \tau_t =
    \left(
    \mathbf{s}_t,\mathbf{a}_t,r_t,d_t,\mathbf{s}_{t+1}
    \right),
\end{equation}
where $d_t$ is the termination flag. In practice, the autoencoder feature $\mathbf{z}_t^{q}$ and perception ground truth $\mathbf{G}_{t}$ are also saved in $\tau_t$, but we omit them for brevity. The mode of each transition is denoted as $m_t\in\{\text{rl},\text{human},\text{rl\_corr},\text{human\_corr}\}$. Rewards include sparse and dense terms. Sparse rewards evaluate parking results, rewarding successful parking and penalizing boundary violation, timeout, and collision. Dense terms deliver stepwise feedback to align the vehicle with the target slot and avoid inefficient or risky motions. Combined outcome assessment and behavioral guidance enable safe and steady parking. Reward formulas and weights are detailed in the Appendix A. The simulator terminates an episode with status $\sigma_t\in\{\text{arrived},\text{collision},\text{timeout},\text{oob}\}$, where $\text{oob}$ denotes out-of-boundary termination, and a failed autonomous segment is detected by
\begin{equation}
    f_t =
    \mathbb{I}
    \left[
    d_t=1\;\land\;
    \sigma_t\in\{\text{collision},\text{timeout},\text{oob}\}
    \right].
\end{equation}

At the beginning of an episode and whenever control is handed back from the human to the RL policy, the scheduler stores the current snapshot as rollback point
\begin{equation}
    \boldsymbol{\xi}_c =
    \left(
    \mathbf{q}_c,t_c,\mathbf{s}_c,n_c,\ell_c
    \right),
\end{equation}
where $\mathbf{q}_c=(x_c,y_c,\psi_c)$ is the vehicle pose, $t_c$ is the simulator time, $\mathbf{s}_c$ is the observation, $n_c$ is the step index, $\ell_c$ is the starting index of the pending RL transition buffer. During normal autonomous execution, RL transitions are not immediately committed to replay memory. Instead, they are temporarily stored in an episode buffer
\begin{equation}
    \mathcal{T}^{rl}_{\text{pend}}
    =
    \{\tau_{\ell_c,rl},\tau_{\ell_c+1,rl},\ldots,\tau_{t,rl}\}.
\end{equation}
If the policy fails after the checkpoint and the segment length satisfies $|\mathcal{T}^{rl}_{\text{pend}}|>N_{\text{min}}$, the failed autonomous rollout is extracted as
\begin{equation}
    \mathcal{T}_{k}^{-}
    =
    \{\tau_{i,rl}\mid i=\ell_c,\ldots,t_f\},
    \quad
    f_{t_f}=1,
\end{equation}
where $k$ denotes the current training episode. The simulator is then restored by the rollback operator
\begin{equation}
    \mathcal{R}(\boldsymbol{\xi}_c):
    (\mathbf{q}_t,t,\mathbf{s}_t,n_t,\ell_t)
    \leftarrow
    \boldsymbol{\xi}_c.
\end{equation}

After rollback, correction starts in human mode, while the operator may switch between human control and RL control during the correction attempt. Human keyboard commands are not stored as raw key states. Instead, the executed motion is converted into a control action through inverse kinematics,
\begin{equation}
    \tilde{\mathbf{a}}_{t,h}^{h}
    =
    \Phi_{\text{ik}}
    \left(
    \mathbf{q}_t,\mathbf{q}_{t+1};L,\Delta t
    \right)
    =
    [\tilde{\delta}_t,\tilde{v}_t],
\end{equation}
where $L$ is the wheelbase and $\Delta t$ is the simulator step interval. This makes human corrections share the same action space as the actor policy. The correction trajectory is decomposed as
\begin{align}
    \mathcal{T}_{k}^{+}
    &=
    \mathcal{T}_{k,h}^{+}
    \cup
    \mathcal{T}_{k,rl}^{+},\\
    \mathcal{T}_{k,h}^{+}
    &=
    \{\tau_{i,h}^{fix}\},
    \quad
    \mathcal{T}_{k,rl}^{+}
    =
   \{\tau_{i,rl}^{fix}\}.
\end{align}
Only successful corrections are written to correction replay memory:
\begin{equation}
    A_k =
    \mathbb{I}
    \left[
    \sigma_{\text{end}}=\text{arrived}
    \;\land\;
    |\mathcal{T}_{k}^{+}|>0
    \right].
\end{equation}
If $A_k=0$, the operator can retry from the same snapshot or discard the failed attempt. This conservative acceptance rule avoids polluting the replay buffer with incomplete human corrections.

The replay memory is organized as a multi-level mistake notebook. Let $\mathcal{B}^{rl}_{N}$ and $\mathcal{B}^{h}_{N}$ be the normal RL and normal human buffers. For each accepted correction episode $k$, CIL-RL creates an episode-level correction region
\begin{equation}
    \mathcal{B}^{corr}_{k}
    =
    \left\{
    \mathcal{B}^{fail}_{k,rl},
    \mathcal{B}^{fix}_{k,rl},
    \mathcal{B}^{fix}_{k,h}
    \right\}.
\end{equation}
The failed autonomous rollout and its successful correction are inserted as
\begin{align}
    \mathcal{B}^{fail}_{k,rl}
    &\leftarrow
    \mathcal{B}^{fail}_{k,rl}
    \cup
    \mathcal{T}_{k}^{-},\\
    \mathcal{B}^{fix}_{k,h}
    &\leftarrow
    \mathcal{B}^{fix}_{k,h}
    \cup
    \mathcal{T}_{k,h}^{+},\\
    \mathcal{B}^{fix}_{k,rl}
    &\leftarrow
    \mathcal{B}^{fix}_{k,rl}
    \cup
    \mathcal{T}_{k,rl}^{+}.
\end{align}
Thus, $\mathcal{B}^{fail}_{k,rl}$ stores what the policy did before failure, while $\mathcal{B}^{fix}_{k,h}\cup\mathcal{B}^{fix}_{k,rl}$ stores how the same local situation was repaired after rollback. The complete training replay set is
\begin{equation}
    \mathcal{D}
    =
    \mathcal{B}^{rl}_{N}
    \cup
    \mathcal{B}^{h}_{N}
    \cup
    \bigcup_{k}
    \mathcal{B}^{corr}_{k}.
\end{equation}

Mini-batches are sampled as paired evidence. Define the set of valid sampling regions as
\begin{equation}
    \Omega =
    \{normal\}
    \cup
    \left\{
    k
    \mid
    |\mathcal{B}^{fail}_{k,rl}|>0,\;
    |\mathcal{B}^{fix}_{k,rl}|+|\mathcal{B}^{fix}_{k,h}|>0
    \right\}.
\end{equation}
The weight of the normal region and the weight of correction region $k$ are
\begin{align}
    w_{normal}
    &=
    |\mathcal{B}^{rl}_{N}|+
    |\mathcal{B}^{h}_{N}|,\\
    w_k
    &=
    |\mathcal{B}^{fail}_{k,rl}|+
    |\mathcal{B}^{fix}_{k,rl}|+
    |\mathcal{B}^{fix}_{k,h}|.
\end{align}
At each pair-sampling step, a region $o\in\Omega$ is selected according to
\begin{equation}
    P(o)=
    \frac{w_o}{\sum_{\bar{o}\in\Omega}w_{\bar{o}}}.
\end{equation}
If $o=normal$, the sampler draws one transition from $\mathcal{B}^{rl}_{N}$ and one from $\mathcal{B}^{h}_{N}$:
\begin{equation}
    (\tau^{rl},\tau^{h})
    \sim
    \mathcal{U}(\mathcal{B}^{rl}_{N})
    \times
    \mathcal{U}(\mathcal{B}^{h}_{N}).
\end{equation}
If $o=k$, the sampler draws a failure-correction pair:
\begin{equation}
    (\tau^{-},\tau^{+})
    \sim
    \mathcal{U}(\mathcal{B}^{fail}_{k,rl})
    \times
    \mathcal{U}
    \left(
    \mathcal{B}^{fix}_{k,rl}
    \cup
    \mathcal{B}^{fix}_{k,h}
    \right).
\end{equation}
For a batch size $B$, the batch is assembled from $B/2$ such pairs,
\begin{equation}
    \mathcal{B}
    =
    \left\{
    \tau_{2j-1},\tau_{2j}
    \right\}_{j=1}^{B/2},
\end{equation}
so that the critic and actor repeatedly see both erroneous behavior and its corrected counterpart. If no valid human or correction data exist, the sampler falls back to standard off-policy RL sampling from $\mathcal{B}^{rl}_{N}$.

The update objective remains the SAC objective \cite{sac}; CIL-RL changes the replay distribution rather than the Bellman equation. For each sampled transition, the next action is drawn as $\mathbf{a}'_{t+1}\sim\pi_{\theta}(\cdot|\mathbf{z}_{t+1}^{\pi})$, and the soft target is
\begin{align}
    y_t =
    r_t+\gamma(1-d_t)
    \big[
    &\min_{i=1,2}
    Q_{\bar{\omega}_i}
    (\mathbf{z}_{t+1}^{q},\phi_a(\mathbf{a}'_{t+1}))
    \nonumber\\
    &-
    \alpha
    \log
    \pi_{\theta}
    (\mathbf{a}'_{t+1}|\mathbf{z}_{t+1}^{\pi})
    \big].
\end{align}
The twin critics are optimized by
\begin{equation}
    J_Q(\omega_i)
    =
    \mathbb{E}_{\tau_t\sim\mathcal{B}}
    \left[
    \left(
    Q_{\omega_i}
    (\mathbf{z}_{t}^{q},\phi_a(\mathbf{a}_{t}))
    -
    y_t
    \right)^2
    \right],
\end{equation}
and the actor is updated by
\begin{align}
    J_{\pi}(\theta)
    =
    \mathbb{E}_{\mathbf{s}_t\sim\mathcal{B},\mathbf{a}_t^{\pi}\sim\pi_{\theta}}
    \left[
    \alpha
    \log\pi_{\theta}(\mathbf{a}_t^{\pi}|\mathbf{z}_t^{\pi})
    \right.
    \nonumber\\
    \left.
    -
    \min_{i=1,2}
    Q_{\omega_i}
    (\mathbf{z}_{t}^{q},\phi_a(\mathbf{a}_{t}^{\pi}))
    \right].
\end{align}
For the privileged critic branch, the auxiliary reconstruction loss is
\begin{equation}
    J_{AE}(\psi)
    =
    \mathbb{E}_{\tau_t\sim\mathcal{B}}
    \left[
    \left\|
    \mathbf{G}_t
    -
    D_{\psi}(E_{\psi}(\mathbf{G}_t))
    \right\|_1
    \right],
\end{equation}
and the overall optimization can be written as
\begin{equation}
    J =
    \sum_{i=1}^{2}J_Q(\omega_i)
    +
    J_{\pi}(\theta)
    +
    \lambda_{AE}J_{AE}(\psi).
\end{equation}
The policy entropy temperature $\alpha$ follows the standard SAC temperature update with target entropy $\bar{\mathcal{H}}$:
\begin{equation}
    J_{\alpha}(\alpha)
    =
    \mathbb{E}_{\mathbf{a}_t^{\pi}\sim\pi_{\theta}}
    \left[
    -\alpha
    \left(
    \log\pi_{\theta}
    (\mathbf{a}_t^{\pi}|\mathbf{z}_t^{\pi})
    +
    \bar{\mathcal{H}}
    \right)
    \right].
\end{equation}

\begin{table}[t]
\centering
\label{alg:cil_serl}
\small
\renewcommand{\arraystretch}{1.05}
\begin{tabular}{@{}r@{\hspace{0.35em}}>{\raggedright\arraybackslash}p{0.96\columnwidth}@{}}
\toprule
\multicolumn{2}{@{}l@{}}{\textbf{Algorithm 1: CIL-SERL}}\\
\midrule
\textbf{1} & $\mathcal{B}^{rl}_{N},\mathcal{B}^{h}_{N},\{\mathcal{B}^{corr}_{k}\}\leftarrow\emptyset$;\\
\textbf{2}  & \textbf{for} episode $k=1,2,\ldots$ \textbf{do}\\
\textbf{3}  & \qquad $\mathcal{T}^{rl}_{\text{pend}},\mathcal{T}^{h}_{\text{pend}}\leftarrow\emptyset$;\\
\textbf{4}  & \qquad $\boldsymbol{\xi}_c\leftarrow(\mathbf{q}_t,t,\mathbf{s}_t,n_t,\ell_t)$;\\
\textbf{5}  & \qquad \textbf{while} $d_t=0$ \textbf{do}\\
\textbf{6}  & \qquad\qquad \textbf{if} $m_t=\text{rl}$ \textbf{then}\\
\textbf{7}  & \qquad\qquad\qquad $\mathbf{a}_t\sim\pi_{\theta}(\cdot|\mathbf{z}_t^{\pi})$; $\tau_t\leftarrow\text{Step}(\mathbf{a}_t)$;\\
\textbf{8}  & \qquad\qquad\qquad $\mathcal{T}^{rl}_{\text{pend}}\leftarrow\mathcal{T}^{rl}_{\text{pend}}\cup\{\tau_t\}$;\\
\textbf{9}  & \qquad\qquad \textbf{if} $m_t=\text{human}$ and not HandBack \textbf{then}\\
\textbf{10}  & \qquad\qquad\qquad $\mathbf{u}_t\leftarrow\text{Keyboard}()$; $\tau_t\leftarrow\text{Step}(\mathbf{u}_t)$;\\
\textbf{11}  & \qquad\qquad\qquad $\tilde{\mathbf{a}}_t^{h}\leftarrow\Phi_{\text{ik}}(\mathbf{q}_t,\mathbf{q}_{t+1})$;\\
\textbf{12}  & \qquad\qquad\qquad $\mathcal{T}^{h}_{\text{pend}}\leftarrow\mathcal{T}^{h}_{\text{pend}}\cup\{(\mathbf{s}_t,\tilde{\mathbf{a}}_t^{h},r_t,d_t,\mathbf{s}_{t+1})\}$;\\
\textbf{13}  & \qquad\qquad \textbf{else if} $\text{HandBack}$ \textbf{then} \\
\textbf{14}  & \qquad\qquad\qquad $\mathcal{B}^{h}_{N}\leftarrow\mathcal{B}^{h}_{N}\cup\mathcal{T}^{h}_{\text{pend}}$;\\
\textbf{15}  & \qquad\qquad\qquad $\boldsymbol{\xi}_s\leftarrow(\mathbf{q}_t,t,\mathbf{s}_t,n_t,\ell_t)$;\\
\textbf{16}  & \qquad\qquad $\mathcal{B}\sim\text{PairSample}(\mathcal{D})$; \\ 
\textbf{17}  & \qquad\qquad $\theta,\omega,\psi,\alpha\leftarrow\text{SACUpdate}(\mathcal{B})$;\\
\textbf{18}  & \qquad \textbf{end while}\\
\textbf{19} & \qquad \textbf{if} $\sigma_{\text{end}}=\text{arrived}$ \textbf{then} \\
\textbf{20} & \qquad\qquad $\mathcal{B}^{rl}_{N}\leftarrow\mathcal{B}^{rl}_{N}\cup\mathcal{T}^{rl}_{\text{pend}}$;\\
\textbf{21} & \qquad \textbf{else if} $|\mathcal{T}^{rl}_{\text{pend}}|>N_{\text{min}}$ \textbf{then}\\
\textbf{22} & \qquad\qquad $\mathcal{T}_{k}^{-}\leftarrow\mathcal{T}^{rl}_{\text{pend}}[\ell_c:t_f]$; \\
\textbf{23} & \qquad\qquad  $\mathcal{R}(\boldsymbol{\xi}_c):(\mathbf{q}_t,t,\mathbf{s}_t,n_t,\ell_t)\leftarrow\boldsymbol{\xi}_c$;\\
\textbf{24} & \qquad\qquad $\mathcal{T}_{k}^{+}\leftarrow\text{Correct}(\pi_{\theta},\text{Human})$;\\
\textbf{25} & \qquad\qquad \textbf{if} $\sigma_{\text{end}}(\mathcal{T}_{k}^{+})=\text{arrived}$ \textbf{then}\\
\textbf{26} & \qquad\qquad\qquad $\mathcal{B}^{corr}_{k}\leftarrow\{\mathcal{T}_{k}^{-},\mathcal{T}_{k,h}^{+},\mathcal{T}_{k,rl}^{+}\}$;\\
\textbf{27} & \qquad\qquad \textbf{else} $\text{RetryOrDiscard}()$;\\
\textbf{28} & \qquad \textbf{else} $\mathcal{T}^{rl}_{\text{pend}}\leftarrow\emptyset$;\\
\textbf{29} & \qquad $\mathcal{D}=\mathcal{B}^{rl}_{N}\cup\mathcal{B}^{h}_{N}\cup\bigcup_{k}\mathcal{B}^{corr}_{k}$;\\
\textbf{30} & \textbf{end for}\\
\bottomrule
\end{tabular}
\end{table}

\section{Experiments}

\subsection{Simulation Experiment}

ParkingWorld is initialized from pretrained REAP-SAC weights and subsequently trained with the correction-in-the-loop mechanism. The human operator monitors the parking process in a bird's-eye-view interface and intervenes using a mouse to adjust speed and steering magnitude and a keyboard to switch directions (see Appendix B). By default, we use a batch size of 32 on an NVIDIA RTX 4090 GPU. Training is also feasible on an NVIDIA RTX 4060 with batch size 8, but performance decreases. Both training and testing are conducted on five reconstructed 3DGS parking scenes containing 239 standard parking slots.

After training, we evaluate ParkingWorld in the proposed 3DGS simulation environment and compare it with two groups of autonomous parking baselines: rule-based planners and camera-based end-to-end methods. For each trial, the target slot is sampled from the standard slots, and the vehicle is initialized in a collision-free feasible sector centered at the target slot, with a 9 m radius, a $\pm 90^\circ$ angular range around the slot direction, and a random yaw from $[-\pi,\pi]$. Each method is evaluated over 200 trials. By default, all experiments use the mean action predicted by the policy network instead of sampling from the action distribution, which is a stricter evaluation setting than that used in REAP \cite{reap}. We use parking success rate (PSR), parking collision rate (PCR), parking timeout rate (PTR), parking boundary-crossing rate (PBR), and number of gear shifts (NGS), where higher PSR and lower PCR, PTR, PBR, and NGS are preferred.

\begin{table}[t]
\centering
\caption{Simulation comparison with autonomous parking baselines on standard slots.}
\label{tab:simulation_results}
\scriptsize
\setlength{\tabcolsep}{2.2pt}
\renewcommand{\arraystretch}{1.08}
\resizebox{\columnwidth}{!}{%
\begin{tabular}{c|c|ccccc}
\toprule
\makecell[c]{\textbf{Type}} & \makecell[c]{\textbf{Methods}} & \makecell[c]{\textbf{PSR}\\(\%) $\uparrow$} & \makecell[c]{\textbf{PCR}\\(\%) $\downarrow$} & \makecell[c]{\textbf{PTR}\\(\%) $\downarrow$} & \makecell[c]{\textbf{PBR}\\(\%) $\downarrow$} & \makecell[c]{\textbf{NGS}\\$\downarrow$} \\
\midrule
\multirow{2}{*}{\makecell[c]{Rule-based\\Planner}} & RS Curve & 31.0 & 0.0 & 69.0 & 0.0 & 1.6 \\
& Hybrid A* & 55.5 & 0.0 & 44.5 & 0.0 & 14.9 \\
\midrule
\multirow{4}{*}{\makecell[c]{End-to-end}} & ParkingE2E & 34.0 & 46.5 & 0.0 & 19.5 & 2.6 \\
& REAP-SAC & 68.5 & 22.0 & 9.5 & 0.0 & 17.5 \\
& ParkingHIL & 72.0 & 19.0 & 9.0 & 0.0 & 16.1 \\
& ParkingWorld & \textbf{88.0} & 4.0 & 8.0 & 0.0 & 13.2 \\
\bottomrule
\end{tabular}
}
\vspace{-1.5em}
\end{table}

\begin{figure}[t!]
   \centering
   \includegraphics[width=0.96\columnwidth]{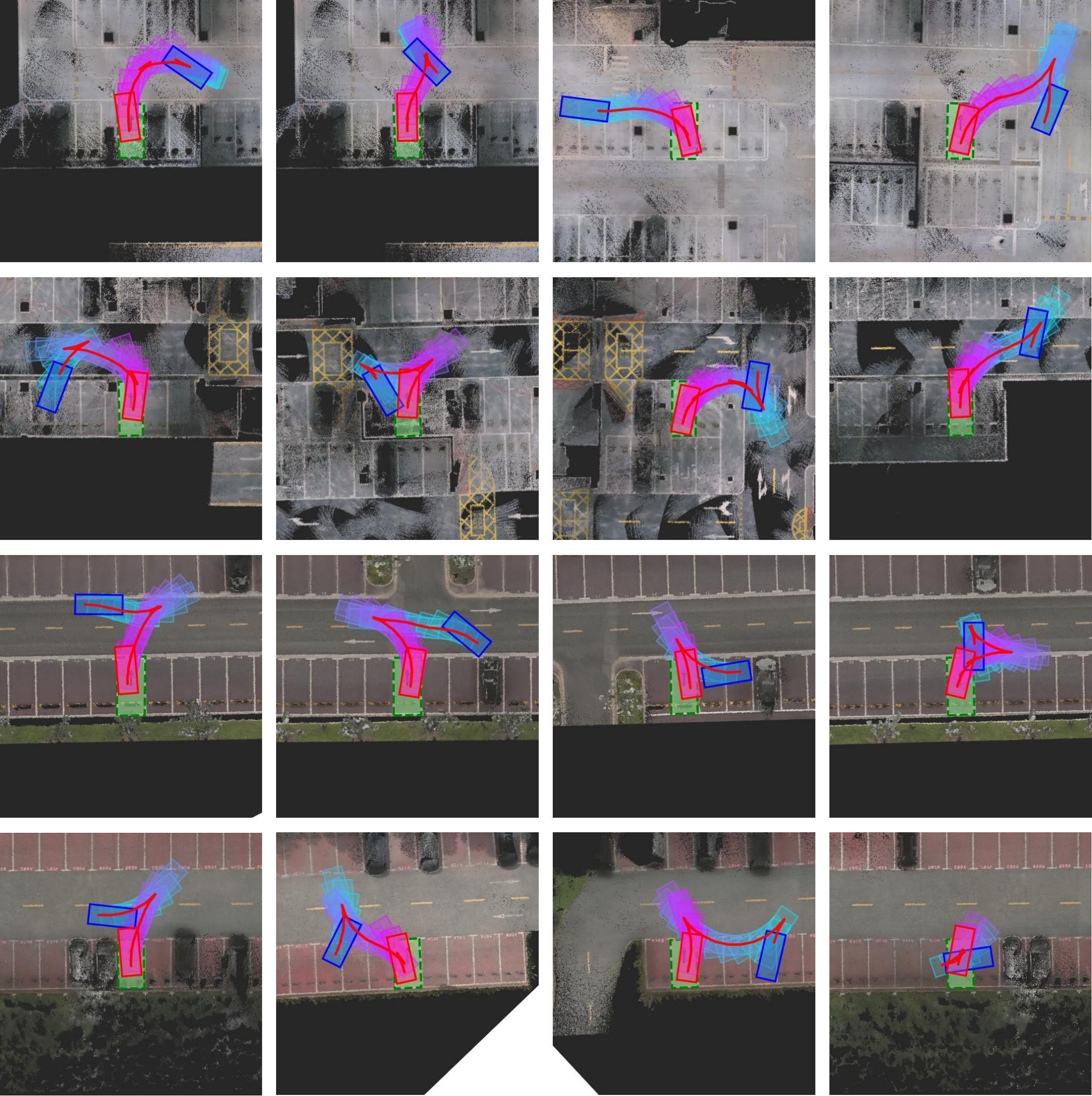}
   \caption{Parking trajectory visualization. The blue box denotes the initial vehicle position, the red box denotes the final vehicle position, and the trajectory of the vehicle rear-axle center is shown in red. Each row corresponds to a different scene.}
   \label{parking_trajectory_visualization}
\end{figure}

Some qualitative results of our method are illustrated in Fig. \ref{parking_trajectory_visualization}. As shown in Table~\ref{tab:simulation_results}, ParkingWorld achieves the best overall performance among all compared methods. Rule-based planners avoid collisions and boundary crossing, but their success rates are limited and they suffer from high timeout rates (solution failure), indicating that purely geometric search is not sufficiently robust in constrained parking scenes. Among end-to-end methods, ParkingWorld achieves the highest PSR of 88.0\%, outperforming the previously best baseline. It also outperforms ParkingHIL, which trains the policy with replay buffers organized following the HIL-SERL \cite{hil-serl} scheme. These results show that learning from corrected failures improves parking safety, reliability, and efficiency in closed-loop simulation.

\subsection{Real-World Experiment}

\begin{figure}[t!]
   \centering
   \includegraphics[width=0.96\columnwidth]{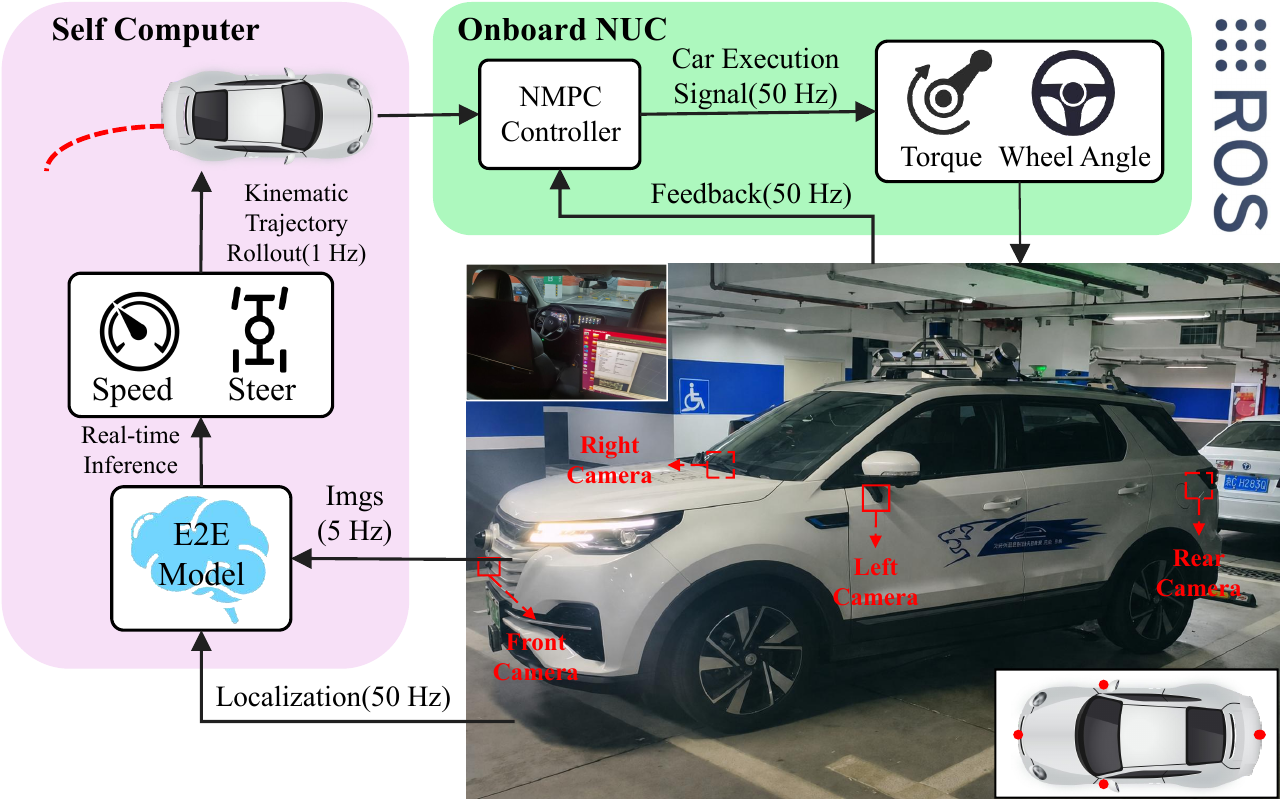}
   \vspace*{0.5em}
   \caption{Real-vehicle deployment framework.}
   \label{real_car_control}
\end{figure}

We deploy ParkingWorld on a real Changan CS55 platform, as shown in Fig.~\ref{real_car_control}. Four surround-view fisheye cameras stream images to a dedicated computer at 5 Hz, where the end-to-end model predicts target speed and steering commands. These commands are rolled out with the same kinematic model used in simulation to generate a short-horizon trajectory, which is refreshed at 1 Hz and sent to the onboard NUC. The onboard NUC uses a sampled NMPC controller to track the predicted trajectory with localization, speed, and steering feedback. It evaluates candidate controls through a kinematic bicycle model and publishes low-level torque, brake, gear, and steering commands at 50 Hz. This hierarchy allows the learned policy to provide parking intent while NMPC compensates for real-vehicle dynamics and execution delay.

\begin{table}[t]
\centering
\caption{Real-vehicle performance comparison of different autonomous parking methods.}
\label{tab:real_world_results}
\scriptsize
\setlength{\tabcolsep}{2.2pt}
\renewcommand{\arraystretch}{1.08}
\resizebox{\columnwidth}{!}{%
\begin{tabular}{c|c|ccccc}
\toprule
\makecell[c]{\textbf{Type}} & \makecell[c]{\textbf{Methods}} & \makecell[c]{\textbf{PSR}\\(\%) $\uparrow$} & \makecell[c]{\textbf{PCR}\\(\%) $\downarrow$} & \makecell[c]{\textbf{PTR}\\(\%) $\downarrow$} & \makecell[c]{\textbf{PBR}\\(\%) $\downarrow$} & \makecell[c]{\textbf{NGS}\\$\downarrow$} \\
\midrule
\multirow{2}{*}{\makecell[c]{Rule-based\\Planner}} & RS Curve & 15.0 & 0.0 & 85.0 & 0.0 & 6.5 \\
& Hybrid A* & 35.0 & 0.0 & 65.0 & 0.0 & 29.7 \\
\midrule
\multirow{4}{*}{\makecell[c]{End-to-end}} & ParkingE2E & 30.0 & 45.0 & 0.0 & 25.0 & 8.5 \\
& REAP-SAC & 55.0 & 15.0 & 30.0 & 0.0 & 27.6 \\
& ParkingHIL & 65.0 & 10.0 & 25.0 & 0.0 & 22.8 \\
& ParkingWorld & \textbf{80.0} & 5.0 & 15.0 & 0.0 & 20.1 \\
\bottomrule
\end{tabular}
}
\vspace{0.5em}

\parbox{\columnwidth}{\scriptsize
\textit{Note:} PCR denotes the driver-observed near-collision rate, determined by the distance to surrounding obstacles, rather than the rate of actual physical collisions.
}
\vspace{-1.8em}
\end{table}

Table~\ref{tab:real_world_results} summarizes the real-vehicle results averaged over 20 trials for each method. Rule-based planners lead to low parking success rates on the physical platform. End-to-end methods better adapt to perception-driven closed-loop parking, but earlier models either collide frequently or fail to complete parking before timeout. ParkingWorld achieves the highest real-vehicle PSR of 80\% and reduces both collision and timeout rates. These results show that correction-in-the-loop fine-tuning improves not only simulation performance but also real-world robustness and safety.

\section{Conclusions And Future Work}
This paper presented ParkingWorld, a correction-in-the-loop sample-efficient reinforcement learning framework for end-to-end autonomous parking in photorealistic 3DGS simulation. By combining a camera-based BEV perception-policy network with structured failure-and-correction replay, the proposed CIL-SERL mechanism enables the policy to learn not only from successful rollouts but also from corrected mistakes. Experiments on five reconstructed parking scenes and real-vehicle deployment on a Changan CS55 demonstrate that ParkingWorld improves parking success rate, reduces unsafe failures, and transfers effectively from 3DGS simulation to the physical platform.

In future work, we plan to combine reinforcement learning with a World-Action-Model that jointly predicts how the surrounding scene and vehicle state evolve under candidate actions. Such a model could provide richer imagination-based rollouts, improve long-horizon credit assignment, and further reduce the amount of human correction needed for robust autonomous parking.

\section*{Appendix}
\setcounter{equation}{0}
\renewcommand{\theequation}{A\arabic{equation}}
\setcounter{table}{0}
\renewcommand{\thetable}{A\arabic{table}}

\subsection{Reward Design}
The reward function combines sparse terminal feedback with dense stepwise guidance. Sparse terms define the task outcome, including successful parking, collision, timeout, and boundary violation. Dense terms refine the continuous parking behavior by encouraging geometric alignment with the target slot, discouraging near-obstacle motions, and penalizing inefficient actions.

We first compute a normalized Euclidean signed distance field (ESDF) from the obstacle set $\mathcal{O}$. Let $d_0=1$ m denote the safety distance and let $\mathbf{x}$ be a point in the workspace. The ESDF value is
\begin{equation}
\text{ESDF}(\mathbf{x})=
\begin{cases}
\dfrac{\min\left(d_0,\min_{\mathbf{y}\in\mathcal{O}}\|\mathbf{x}-\mathbf{y}\|\right)}{d_0}, & \mathbf{x}\notin\mathcal{O},\\[0.6em]
0, & \mathbf{x}\in\mathcal{O}.
\end{cases}
\end{equation}
Thus, $\text{ESDF}(\mathbf{x})\in[0,1]$, where larger values indicate safer clearance.

The successful parking condition is evaluated by the overlap and orientation consistency between the vehicle footprint $A_t$ and the target slot polygon $B$. The overlap score is
\begin{equation}
\text{IoU}_t=\frac{S(A_t\cap B)}{\min(S(A_t),S(B))},
\end{equation}
where $S(\cdot)$ denotes polygon area. A success reward is given when the vehicle satisfies the arrival condition, i.e., $\text{IoU}_t>0.9$ and the heading error $\Delta\theta_t<75^\circ$:
\begin{equation}
r_{\text{success}}=
\begin{cases}
c_{\text{success}}, & \text{arrived},\\
0, & \text{otherwise}.
\end{cases}
\end{equation}

To provide dense alignment feedback, we reward progress in vehicle-slot overlap only when the heading error is within the valid range:
\begin{equation}
r_{\text{union}}=
\mathbb{I}\!\left[\Delta\theta_t<75^\circ\right]
w_{\text{union}}
\left(\text{IoU}_t-\max_{1\le j<t}\text{IoU}_j\right).
\end{equation}
This incremental reward encourages the policy to gradually improve the parking pose instead of relying only on terminal success feedback.

Safety is enforced by hard and soft collision penalties. A hard collision penalty is triggered when the vehicle footprint intersects an occupied obstacle region:
\begin{equation}
r_{\text{collision}}=-c_{\text{collision}}.
\end{equation}
The soft collision term acts before an actual collision. For each action, the simulator rolls out $n$ short substeps and obtains intermediate vehicle states $\mathcal{S}_t=\{\mathbf{s}_0,\ldots,\mathbf{s}_{n-1}\}$. Let $\mathcal{B}(\mathbf{s})$ denote the sampled boundary points of the vehicle at state $\mathbf{s}$. The soft collision penalty is
\begin{equation}
r_{\text{soft}}=
w_{\text{soft}}
\left(
\min_{\mathbf{s}\in\mathcal{S}_t}
\min_{\mathbf{x}\in\mathcal{B}(\mathbf{s})}
\text{ESDF}(\mathbf{x})-1
\right).
\end{equation}
Since the ESDF is normalized to $[0,1]$, this term becomes more negative when the predicted short-horizon motion approaches obstacles.

The boundary penalty is applied when the vehicle leaves the valid operation region, which is represented by the distance from the vehicle center to the target slot center:
\begin{equation}
r_{\text{outbound}}=-c_{\text{outbound}}.
\end{equation}
In our implementation, the episode is regarded as out of bounds when this distance exceeds 15 m. To avoid local deadlock, a stuck penalty is applied when the vehicle displacement between two consecutive simulation steps is below $0.01$ m:
\begin{equation}
r_{\text{stuck}}=-c_{\text{stuck}}.
\end{equation}

The time cost encourages efficient parking while avoiding domination over safety terms:
\begin{equation}
r_{\text{time}}=-w_{\text{time}}\tanh\left(\frac{t}{10T_{\text{tol}}}\right),
\end{equation}
where $T_{\text{tol}}=120$ is the maximum tolerated number of parking steps. If the task is not completed within this horizon, a timeout penalty is assigned:
\begin{equation}
r_{\text{outtime}}=-c_{\text{outtime}}.
\end{equation}

The final reward at each step is the weighted sum of all active terms:
\begin{equation}
\begin{aligned}
r_t=w_{\text{reward}}\big(&
r_{\text{success}}+r_{\text{union}}+r_{\text{collision}}
+r_{\text{soft}} \\
&+r_{\text{outbound}}+r_{\text{stuck}}
+r_{\text{time}}+r_{\text{outtime}}
\big).
\end{aligned}
\end{equation}
The reward parameters used in our experiments are listed in Table~\ref{tab:reward_parameters}.

\begin{table}[h]
\centering
\caption{Reward parameters used in ParkingWorld.}
\label{tab:reward_parameters}
\small
\setlength{\tabcolsep}{4pt}
\renewcommand{\arraystretch}{1.05}
\begin{tabular}{c|c|c}
\toprule
\textbf{Symbol} & \textbf{Meaning} & \textbf{Value} \\
\midrule
$c_{\text{success}}$ & successful parking reward & 50 \\
$c_{\text{outbound}}$ & boundary violation penalty & 10 \\
$c_{\text{collision}}$ & hard collision penalty & 50 \\
$c_{\text{stuck}}$ & stuck penalty & 0.3 \\
$c_{\text{outtime}}$ & timeout penalty & 3 \\
$w_{\text{union}}$ & overlap progress weight & 10 \\
$w_{\text{time}}$ & time cost weight & 3 \\
$w_{\text{soft}}$ & soft collision weight & 0.3 \\
$w_{\text{reward}}$ & global reward scale & 0.1 \\
\bottomrule
\end{tabular}
\end{table}

\setcounter{subsection}{1}
\renewcommand{\theHsubsection}{B}
\subsection{Human-in-the-Loop Training Dynamics}
Starting from a purely RL-trained policy, human-in-the-loop training introduces high-value corrective demonstrations around failure-prone states, yielding rapid early improvement. However, continued training may shift the replay distribution toward intervention states. Since our current interface uses mouse and keyboard inputs, human corrections are relatively discrete, which may bias value estimation and cause partial overfitting to correction behaviors. As a result, performance can decrease after reaching a peak, while still remaining better than the original RL policy. If steering wheels and pedals are used to provide a more continuous mapping of human interventions, further performance improvement can be expected.

\end{document}